\documentclass{article}

\usepackage{enumitem}
\usepackage{verbatim}
\usepackage{PRIMEarxiv}
\usepackage{enumitem}
\usepackage{verbatim}
\usepackage{fancyhdr}
\usepackage[utf8]{inputenc} % allow utf-8 input
\usepackage[T1]{fontenc}    % use 8-bit T1 fonts
\usepackage{hyperref}       % hyperlinks
\usepackage{url}            % simple URL typesetting
\usepackage{booktabs}       % professional-quality tables
\usepackage{amsfonts} % blackboard math symbols
\usepackage{nicefrac}       % compact symbols for 1/2, etc.
\usepackage{microtype}      % microtypography
\usepackage{lipsum}
\usepackage{fancyhdr}       % header
\usepackage{graphicx}       % graphics
\usepackage{multirow}
\usepackage{graphicx}
\usepackage{textcomp}
\usepackage[export]{adjustbox}
\graphicspath{{media/}}     % organize your images and other figures under media/ folder

\title{AERoS: Assurance of Emergent Behaviour in Autonomous Robotic Swarms\thanks{\dag D.B.\ Abeywickrama and J.\ Wilson contributed equally.}}

\author{
   Dhaminda B. Abeywickram{a}$^\dag$\\
   Department of Computer Science \\
   University of Bristol, UK \\
   \texttt{dhaminda.abeywickrama@bristol.ac.uk} 
   \And
   James Wilso{n}$^\dag$ \\
	Department of Engineering Mathematics \\
	University of Bristol, UK \\
	\texttt{wilson-james@outlook.com} 
   \And
   Suet Lee \\
	Department of Engineering Mathematics \\
	University of Bristol, UK \\
   \texttt{suet.lee@bristol.ac.uk} 
   \And
    Greg Chance \\
	Department of Computer Science \\
	University of Bristol, UK \\
	\texttt{greg.chance@bristol.ac.uk} 
   \And
   Peter D. Winter \\
   School of Sociology \\
	University of Bristol, UK \\
   \texttt{peter.winter@bristol.ac.uk} 
   \And
   Arianna Manzini \\
   Bristol Medical School\\
	University of Bristol, UK \\
   \texttt{amanzini.ethics@gmail.com} 
   \And    
   Ibrahim Habli \\
   Department of Computer Science \\
   University of York, UK \\
   \texttt{ibrahim.habli@york.ac.uk} 
   \And
   Shane Windsor \\
	Department of Aerospace Engineering \\
	University of Bristol, UK \\
	\texttt{shane.windsor@bristol.ac.uk} 
   \And
   Sabine Hauert \\
	Department of Engineering Mathematics \\
	University of Bristol, UK \\
	\texttt{sabine.hauert@bristol.ac.uk}    
   \And
   Kerstin Eder \\
   Department of Computer Science \\
   University of Bristol, UK \\
   \texttt{kerstin.eder@bristol.ac.uk} 
}

\begin{document}
\maketitle
\begin{abstract}
The behaviours of a swarm are not explicitly engineered. Instead, they are an emergent consequence of the interactions of individual agents with each other and their environment.
This emergent functionality poses a challenge to \emph{safety assurance}. 
The main contribution of this paper is a process for the safety assurance of emergent behaviour in autonomous robotic swarms called \emph{AERoS}, following the guidance on the Assurance of Machine Learning for use in Autonomous Systems (AMLAS). 
We explore our proposed process using a case study centred on a robot swarm operating a public cloakroom.
\end{abstract}
\keywords{Assurance \and Safety \and Emergent Behaviour \and Guidance \and Swarms}
%%------------------------------------------------------------------------------------------------------------------------------------------------------------------------------------------------------------------------------
\newpage
\section{Introduction}\label{introduction}
Swarm robotics provides an approach to the coordination of large numbers of robots inspired by swarm behaviours in nature~\cite{Sahin2005}. 
The overall behaviours of a swarm are not explicitly engineered in the system. Instead, they are an emergent consequence of the interactions of individual agents with each other and the environment~\cite{Abeywickrama2022}; this poses a challenge to \emph{assurance}. 
According to the ISO standard for systems and software engineering vocabulary~\cite{ISO24765:2017}, \emph{assurance} is defined as ``all the planned and systematic activities implemented within the quality system, and demonstrated as needed, to provide adequate confidence that an entity will fulfil requirements for quality''. 
Assurance tasks comprise conformance to standards, verification and validation (V\&V), and certification. Assurance criteria for autonomous systems (AS) include both functional and non-functional requirements such as safety~\cite{Cheng2014}. 

Existing standards and regulations of AS are either implicitly or explicitly based on the V lifecycle model~\cite{Forsberg:1992}, which moves from requirements through design onto implementation and testing before deployment~\cite{Jia2021,Fisher2020}. 
However, this model is unlikely to be suitable for systems with emergent behaviour (EB); for example\ through interaction with other agents and the environment, as is the case with swarms. 
ISO standards have been developed for the service robotics sector (non-industrial) (e.g.\ ISO 13482, ISO 23482-1, ISO 23482-2), and the industrial robotics sector (e.g.\ ISO 10218-1, ISO 10218-2, ISO/TS 15066)~\cite{Abeywickrama2022}. 
However, although these standards focus on ensuring the assurance of robots at the individual level, they do not cover safety or any other extra-functional property at the \emph{swarm} level that may arise through EB. 

The main contribution of this paper is a process for the safety assurance of EB in autonomous robotic swarms (AERoS), adapted from the guidance on the Assurance of Machine Learning for use in Autonomous Systems (AMLAS)~\cite{Hawkins2021}. 
AERoS covers six EB lifecycle stages: safety assurance scoping, safety requirements elicitation, data management, model EB, model verification, and  model deployment. 
The AERoS process is domain independent and therefore can be applied to any swarm type (e.g. grounded, airborne). 
In this paper, we explore it using a case study centered on a robot swarm operating a public cloakroom at events with 50 to 10000 attendees~\cite{Jones2020}. 
In the cloakroom, a swarm of robots assist attendees to deposit, store, and deliver their belongings (e.g. jackets) \cite{Jones2020}. 
As the swarm operates in a public setting, the system must prioritise public safety. 

The rest of the paper is organised as follows. 
In Section~\ref{relatedwork}, we provide key related work to our study. Section~\ref{framework} discusses the six stages of the AERoS process. Finally, Section~\ref{discussion-conclusions} provides a brief discussion and concludes the paper. 

\section{Related Work}\label{relatedwork}
AS are considered to follow a much more iterative life-cycle compared to the conventional V-model. 
Thus, there is a need for new standards and assurance processes that extend beyond design time and allow continuous certification at runtime~\cite{Rushby2008}. 
In this context, there have been several standards and guidance introduced by various industry committees and research groups. In 2016, the British Standards Institution introduced the BS 8611 standard that provides a guide to the ethical design and application of robots and robotic systems. 
Then, IEEE through its Global Initiative on Ethics of Autonomous and Intelligent Systems initiated the development of a series of standards to address autonomy, ethical issues, transparency, data privacy and trustworthiness (e.g. IEEE P7001~\cite{IEEE-P7001}, P7007, P7010).  
There are several standards and guidance related to machine learning in aeronautics, automotive, railway and industrial domains \cite{Kaakai2022}, for example the AMLAS process~\cite{Hawkins2021}, the European Union Aviation Safety Agency (EASA) concept paper, the DEpendable and Explainable Learning (DEEL) white paper, the Aerospace Vehicle System Institute (AVSI) report, the Laboratoire National de Métrologie et d'Essais (LNE) certification, and the UL 4600 standard. However, none of these approaches targets robot swarms. 

In this work we used AMLAS \cite{Hawkins2021} as the foundation for developing an assurance process for autonomous robotic swarms. 
AMLAS provides guidance on how to systematically integrate safety assurance into the development of the machine learning components based on offline supervised learning. 
AMLAS contains six stages where assurance activities are performed in parallel to the development activities. 
AMLAS has the advantage of ensuring safety assurance for complex AS where the behaviour of the system is controlled by machine learning algorithms. 
In this work, we take inspiration from AMLAS, but adapt it to focus on emergence as the driver for complexity, rather than learning. 

\section{The AERoS Process}\label{framework}
\begin{figure*}[!t]
	\centering
	\includegraphics[width=0.85\textwidth]{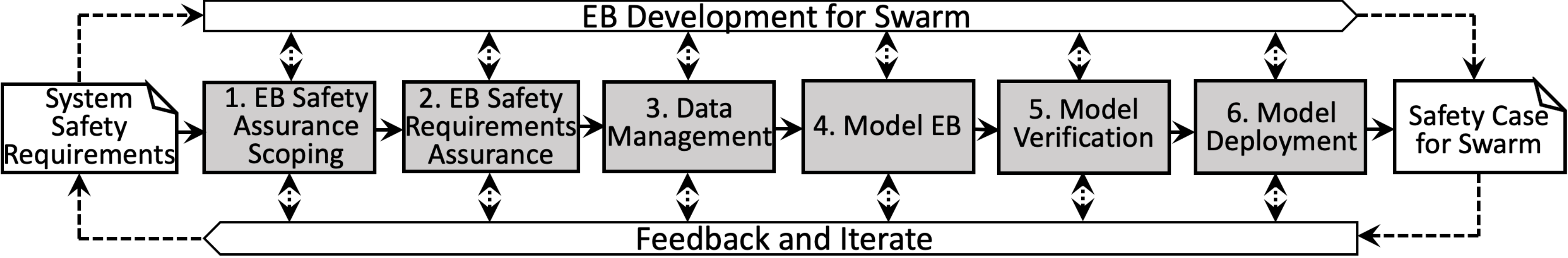}
	\caption{The AERoS process with the six stages adapted from AMLAS.}
	\label{aeros-process}
\end{figure*}

This section discusses the six main stages of AERoS targeting autonomous robot swarms. AERoS is iterative by design, and the assurance activities are performed in parallel to EB development (see Fig.~\ref{aeros-process}). For each stage, we describe its inputs and outputs, main assurance activities and their associated artefacts. 

\subsection{Stage 1: EB Safety Assurance Scoping} \label{framework-stage1}
\begin{figure}[!h]
	\centering
	\centering
	\includegraphics[width=0.45\textwidth]{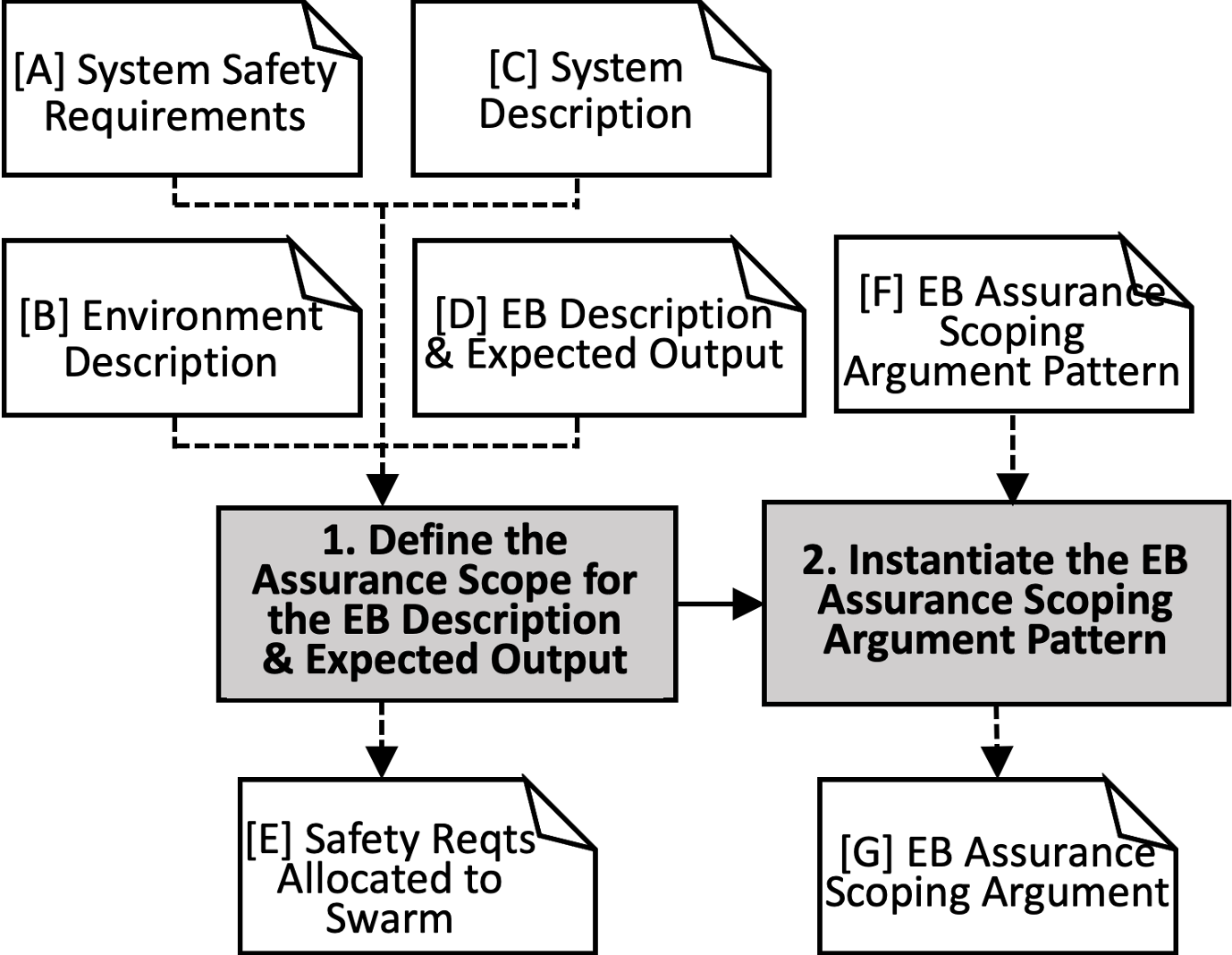}
	\caption{Stage 1: The AERoS emergent behaviour assurance scoping process.}
	\label{aeros-stage1}
\end{figure}

Stage 1 contains two activities which are performed to define the safety assurance scope for the swarm (see Fig.~\ref{aeros-stage1}). 

\noindent\textbf{Activity 1. Define Assurance Scope for the EB Description and Expected Output}
The goal of Activity 1, which has four inputs [A--D] (Fig.~\ref{aeros-stage1}), is to define the safety assurance scope for the EB and expected output. The output of this activity is the Safety Requirements Allocated to the Swarm [E]. The requirements defined in this stage are independent of any EB technology, which reflects the need for the robot swarm to perform safely regardless of emergence.  

\begin{figure}[!h]
	\centering
	\includegraphics[width=0.40\textwidth]{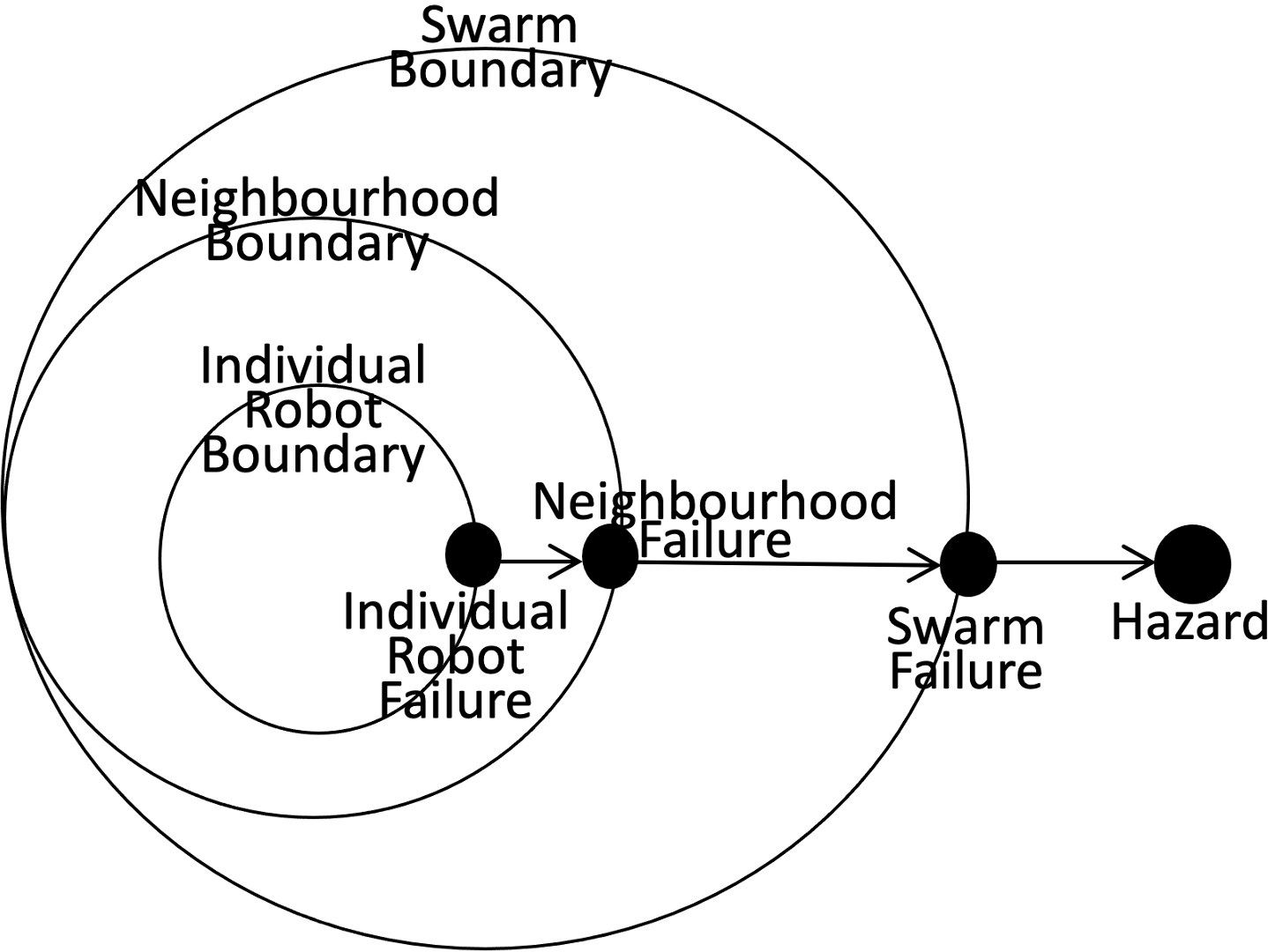}
	\caption{Failure conditions in a swarm adapted from DO-178C and AMLAS.}
	\label{failure-events}
\end{figure}

\emph{[A] System Safety Requirements:}
The system safety assessment process generates the safety requirements of the swarm, and covers the identification of hazards (e.g. the blocking of critical paths in the cloakroom) and risk analysis.
Figure~\ref{failure-events} illustrates how individual robot failures propagate through the neighbourhood to swarm-level hazards: we can then derive safety requirements in the form of concrete failure conditions at the level of the whole swarm which capture, implicitly, all levels of the swarm. 
Although this has been illustrated as a simplified linear chain of events, in reality this represents a complex sequence which can be difficult to distil into distinct events and causes. 

\emph{[B] Environment Description:}
It is essential to consider the system environment when allocating safety requirements to the swarm. 
In the cloakroom, a swarm of robotic agents collects and delivers jackets, which are stored in small box-like containers. 
The agents are required to navigate a public space between collection and delivery points. 

\begin{figure}[!h]
	\centering
	\centering
	\includegraphics[width=0.42\textwidth]{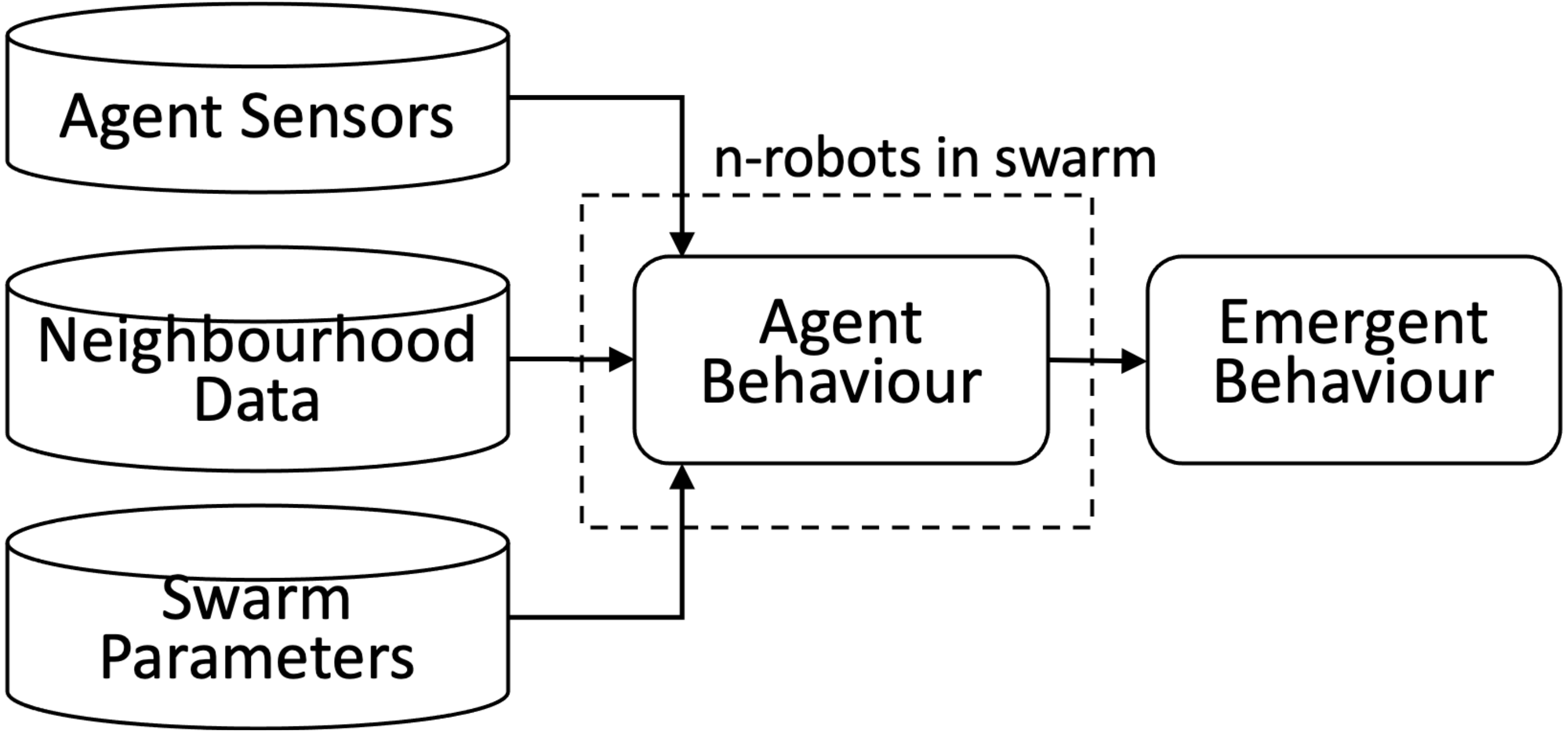}
	\caption{Inputs fed into individual agent behaviour producing overall swarm emergent behaviour. }
	\label{system-description}
\end{figure}

\emph{[C] System Description:}
In the cloakroom, we can consider three inputs: sensor availability, neighbourhood data (used because there is no access to global data in real-world deployments), and swarm parameters (see Fig.~\ref{system-description}). The \emph{sensors} available to agents can be cameras and laser time-of-flight sensors. 
The neighbourhood data of the swarm can be specified through the communication systems available to agents, in this case Bluetooth. 
Through the use of this short-range communication, agents can access neighbourhood data, such as approximate position or current state of local agents.  
As for the swarm-level parameters, we can consider options specified by a user, that is, the number of agents deployed, and the maximum speed of agents. 
Once defined, the three inputs are then fed to the individual agents to instruct their behaviour. This behaviour enacted by multiple agents then produces a swarm-level EB as the individuals interact with one another and their environment. 

\emph{[D] EB Description and Expected Output:}
By expected output, we refer to the gains that can arise from the system by deploying multiple agents. 
In the cloakroom, the output is a collaborative system capable of collecting, sorting, and delivering jackets in a public setting. 
To achieve this, the EB of the system needs to arise from the available behaviours of the individual agents, their interactions, and the constraints outlined in the system description.

\begin{figure}[!h]
		\centering
		\includegraphics[width=0.5\textwidth]{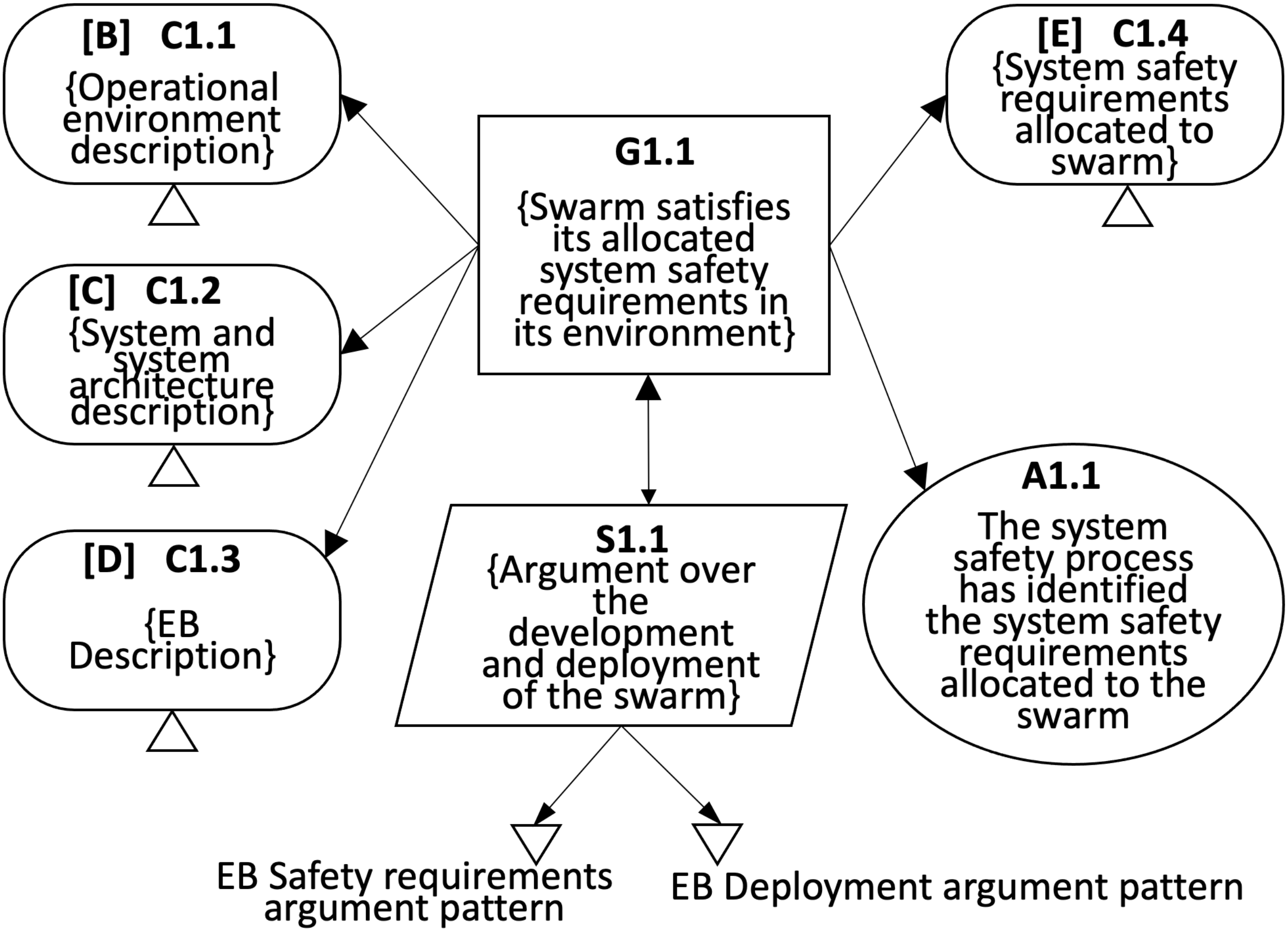}
		\caption{Emergent behaviour safety assurance scoping argument pattern.}
		\label{stage1-ap}
\end{figure}

\noindent\textbf{Activity 2. Instantiate EB Assurance Scoping Argument Pattern} Each stage of the AERoS process includes an activity to instantiate a safety argument pattern based on the evidence and artefacts generated in that stage. 
\emph{Argument patterns}~\cite{Hawkins2021}, which are modelled using the Goal Structuring Notation, can be used to explain the extent to which the evidence supports the relevant EB safety claims.  
In Activity 2, we use the artefacts generated from Stage 1 (i.e. [A–E]) to instantiate the EB Assurance Scoping Argument Pattern ([F] – see Fig.~\ref{stage1-ap}). 
The instantiated argument [G] along with other instantiated arguments resulting from the other five stages of AERoS constitute the safety case for the swarm. The activities to instantiate argument patterns of the other stages follow a very similar pattern so are not shown due to space limitations.
\subsection{Stage 2: EB Safety Requirements Assurance} \label{framework-stage2}
\begin{figure}[!h]
	\centering
	\centering
	\includegraphics[width=0.5\textwidth]{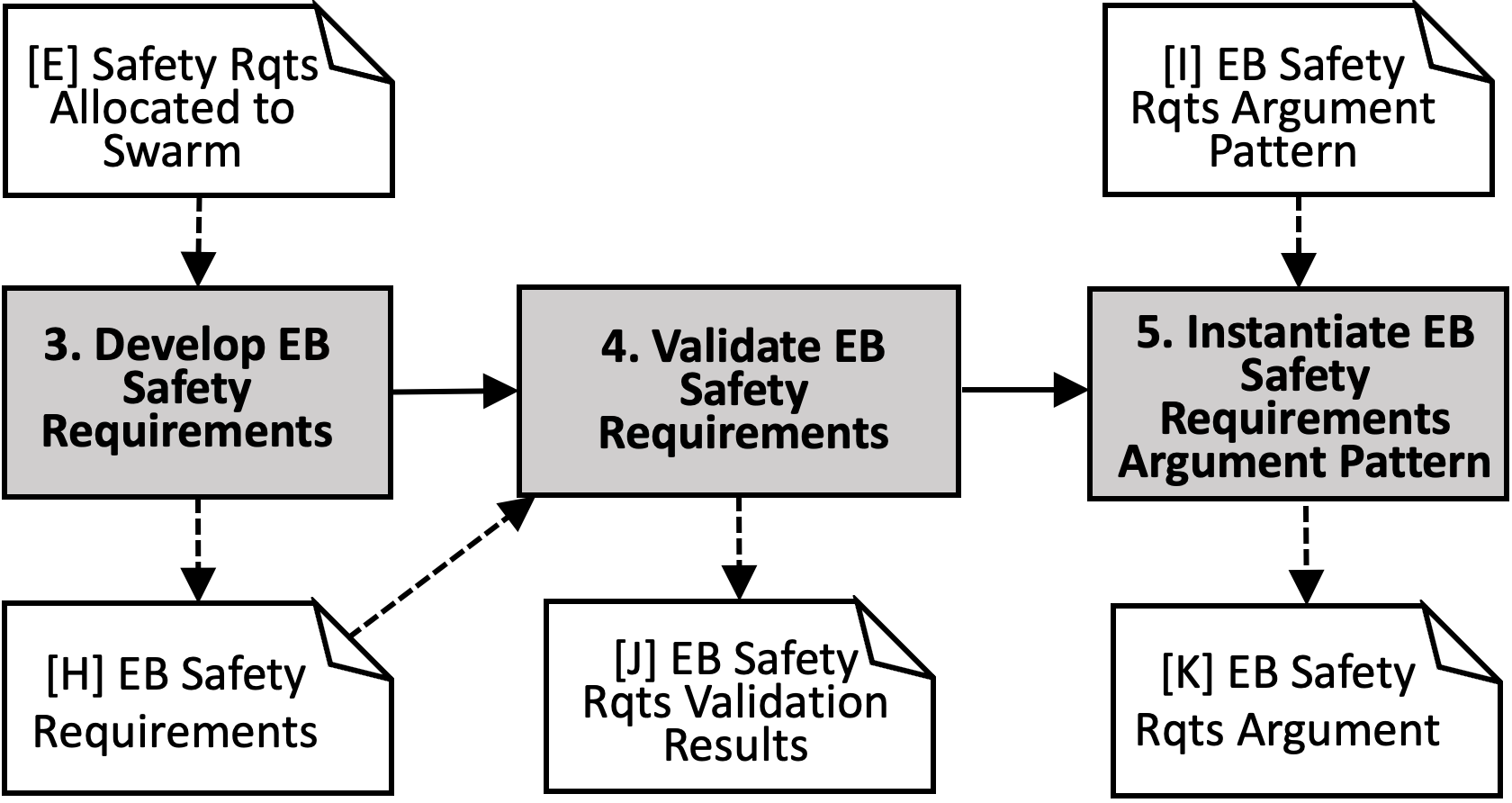}
	\caption{Stage 2: The AERoS emergent behaviour safety requirements assurance.}
	\label{aeros-stage2}
\end{figure}

Stage 2 contains three activities (Fig.~\ref{aeros-stage2}), which are performed to provide assurance in EB safety requirements for the swarm. 

\noindent\textbf{Activity 3. Develop EB Safety Requirements} The input to Activity 3 in Stage 2 is the Safety Requirements Allocated to the Swarm [E]. 
We define EB safety requirements to specify risk controls for the swarm-level hazards by taking into account the system architecture defined and the operating environment. 

In the swarm context, we consider four types of requirements: \emph{performance}, \emph{adaptability}, \emph{human safety}, and \emph{environment}. 
In particular, the environment requirements capture the need for the system to be robust to variation in the operative space.
We consider several safety metrics under each requirement category: 
(i) performance: low-impact and high-impact collisions (the swarm should operate below a \emph{critical number} of such collisions); 
(ii) adaptability: percentage of swarm stationary outside of the delivery site, number of stationary agents, time since last agent moved; 
(iii) human safety: velocity or average velocity of agents, swarm size, rate of humans encountered, proximity to humans;
(iv) environment: sum of objects in an area (the density of objects in the environment should not block swarm operation).
As the robot swarm is composed of many agents, there is potential for a large number of faults to occur at any given time~\cite{Lee2022}. This motivates three further sub-categories for each of the performance, adaptability and human-safety requirements: \emph{faultless operations}, \emph{failure modes (graceful degradation)}, and \emph{worst case}. 
\emph{Graceful degradation} refers to the acceptable level of faults, their impact, and how the system should react when faults occur. \emph{Worst case} accounts for the least acceptable impact the system should experience and the means to avoid it. 
A key output of Activity 3 is [H], which describes the EB Safety Requirements relating to: performance, adaptability, and environment (see Table~\ref{tab:reqs}), as well as human safety (Table~\ref{tab:human-s}). 
These example requirements have been generated following three main considerations: the hazards for each safety requirement type, the metrics~\cite{Lee2022} available to assess these hazards, and the realistic thresholds~\cite{Jones2022} given the specification of the system.

\begin{figure}[!h]
	\centering
	\includegraphics[trim={30mm 25mm 45mm 30mm},clip,width=0.4\textwidth]{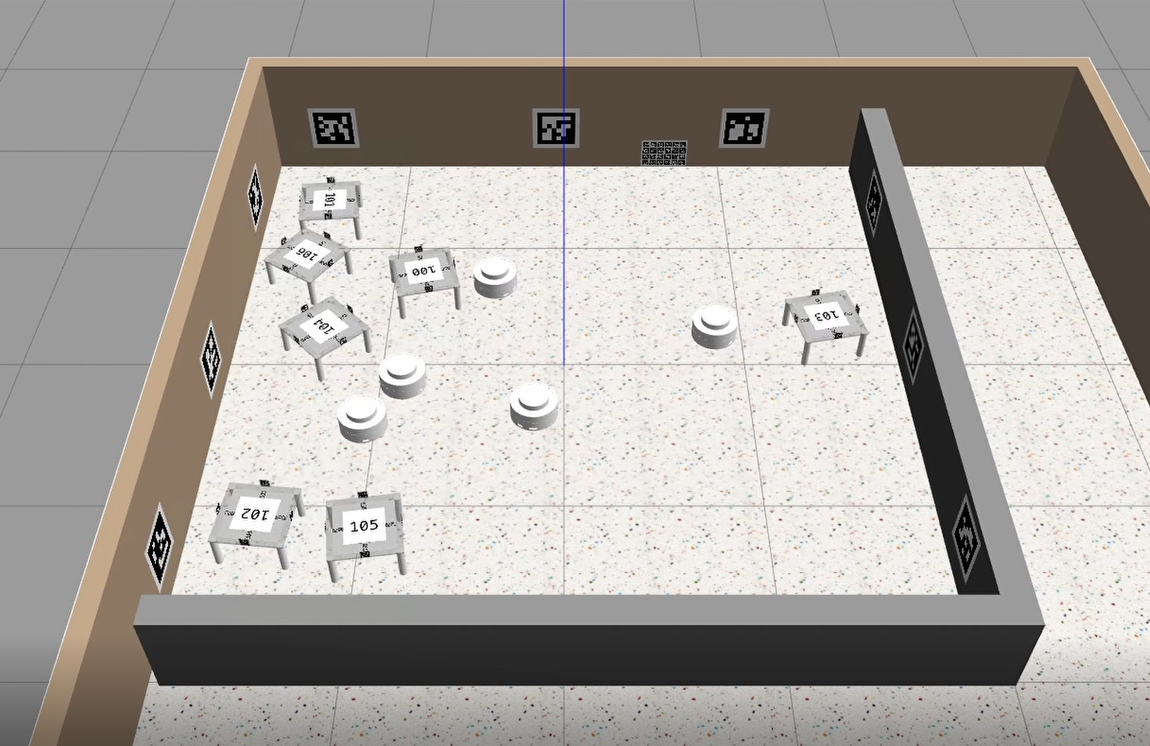}
	\caption{3D simulation created to validate several emergent behaviour safety requirements.}
	\label{3Dsim}
\end{figure}

\begin{table}[!t]
\centering
\caption{\label{tab:reqs}Examples of performance, adaptability, and environmental safety requirements for the cloakroom scenario.}
\begin{tabular}{p{5mm} p{140mm} }
\textbf{RQ} & \textbf{Performance Requirements}\\
\hline
1.1 & The swarm \emph{shall} experience \textbf{$<$ 1 high-impact (V $>$ 0.5m/s)} collisions across \textbf{a day} of \textbf{faultless} operation. \\ 
\hline
1.2 & The swarm \emph{shall} experience \textbf{$<$ 0.1\%} increase in \textbf{high-impact} collisions across \textbf{a day's} operation with \textbf{full communication faults} occurring in \textbf{10\%} of the swarm.\\ 
\hline
1.3 & The swarm \emph{shall} experience \textbf{$<$ 0.1\%} increase in \textbf{high-impact} collisions across \textbf{a day's} operation with \textbf{half-of-wheels motor faults} occurring in \textbf{50\%} of the swarm.	\\	
\hline
1.4 & The swarm \emph{shall} experience \textbf{$<$ 2 high-impact (V $>$ 0.5m/s)} collisions across \textbf{a day} of \textbf{faulty} operation.  \\		 		
\hline
1.5 & The swarm agents \emph{shall} \textbf{weigh $<$ 3kg} and shall have \textbf{acceleration} \textbf{$<$ 4m/s} so that the \textbf{maximum collision force} in the swarm is within acceptable bounds. \\
\hline
1.6 & The swarm agents \emph{shall} only carry objects of \textbf{weight $<$ 2kg}. \\ 
\hline \\[-1.25\medskipamount]
& \textbf{Adaptability Requirements}\\
\hline
2.1 & The swarm \emph{shall} have \textbf{$<$ 10\%} of its agents \textbf{stationary*} outside of the \textbf{delivery site} at a given time.
*Assumption: Agents are considered stationary once they have not moved for $>$ \textbf{10 seconds}.
\\ 
\hline
2.2 & All agents of the swarm \emph{shall} move at least every \textbf{100 seconds} if outside of the \textbf{delivery site}.\\ 
\hline
2.3 & The swarm \emph{shall} experience $<$ \textbf{10\%} increase in the \textbf{number of stationary agents} at any time with \textbf{half-of-wheels motor faults} occurring in \textbf{50\%} of the swarm. \\
\hline
2.4 & The swarm agents \emph{shall} experience $<$ \textbf{10\%} increase in \textbf{stationary time} with \textbf{half-of-wheels motor faults} occurring in \textbf{50\%} of  the swarm.\\ 
\hline
2.5 & The swarm \emph{shall} experience $<$ \textbf{10\%} increase in \textbf{number of stationary agents} at any given time with \textbf{full communication faults} occurring in \textbf{10\%} of the swarm.\\
\hline
2.6 & The swarm agents \emph{shall} experience $<$ \textbf{10\%} increase in \textbf{stationary time} with \textbf{full communication faults} occurring in \textbf{10\%} of the swarm. \\	
\hline
2.7 & The swarm \emph{shall} have \textbf{$<$ 20\%} of its agents \textbf{stationary*} outside of the \textbf{delivery site} at a given time.
*Assumption: Agents are considered stationary once they have not moved for $>$ \textbf{10 seconds}. \\ 
\hline \\[-1.25\medskipamount]
& \textbf{Environmental Requirements} \\ 
\hline
3.1 & The swarm \emph{shall} perform as required in environmental density levels \textbf{0-4 p$_o$ of objects} (sum of boxes and agents per m$^2$) in the environment. 
\\ 
\hline
3.2 & The swarm \emph{shall} perform as required when \textbf{floor incline} is \textbf{0-20 degrees}.
\\ 
\hline
3.3 & The swarm \emph{shall} perform as required in a \textbf{dry environment}.
\\ 
\hline
3.4 & The swarm \emph{shall} perform as required in \textbf{smooth-floored environments} with step increases no greater than \textbf{0.5cm}.
\\ 
\hline
3.5 & The swarm \emph{shall} only operate in \textbf{environments where humans have devices that identify the human’s location} to the swarm agents. 
\\ 		
\hline \\[-1\medskipamount]
\end{tabular}
\end{table}
\begin{table}[!h]
\centering
\caption{\label{tab:human-s}Examples of human-safety requirements for the cloakroom scenario.}
\begin{tabular}{p{6mm} p{140mm}}
\textbf{RQ} & \textbf{Human-Safety Requirements} \\
\hline
4.1 & The swarm agents \emph{shall} travel at speeds of less than \textbf{0.5m/s} when within \textbf{2m} distance of a \textbf{trained human} (a worker who has received relevant training).
\\ 
\hline
4.2 & The swarm agents \emph{shall} travel at speeds of less than \textbf{0.25m/s} when within \textbf{3m} distance of an \textbf{attendee}.
\\ 
\hline
4.3 & The swarm agents \emph{shall} only come within \textbf{2m} distance of a \textbf{human $<$ 10} times collectively across \textbf{1000 seconds} of \textbf{faultless} operations.
\\ 
\hline
4.4 & The swarm \emph{shall} only allow \textbf{$<$ 5 agents} to request intervention from a \textbf{trained human} at a given time.
\\ 
\hline
4.5 & A \textbf{trained human} \emph{shall} monitor \textbf{5-20 agents} at a given time.
\\ 
\hline
4.6 & The swarm \emph{shall} only allow \textbf{1 agent} to request input from an \textbf{attendee} at a given time.
\\ 
\hline
4.7 & An \textbf{attendee} \emph{shall} receive  information from $<$ \textbf{5 agents} of the swarm at a given time.
\\ 
\hline
4.8 & The swarm \emph{shall} experience \textbf{$<$ 10\%} increase in \textbf{human encounters} across \textbf{1000 seconds} of operation with \textbf{full communication faults} occurring in \textbf{10\%} of the swarm. \\
\hline
4.9 & The swarm \emph{shall} experience \textbf{$<$ 10\%} increase in \textbf{human encounters }across \textbf{1000 seconds} of operation with \textbf{half-of-wheels motor faults} occurring in \textbf{50\%} of the swarm.\\
\hline
4.10 & The swarm agents \emph{shall} only come within \textbf{2m} distance of a \textbf{human $<$ 20} times collectively across \textbf{1000 seconds} of \textbf{faulty} operations.
\\				
\hline \\[-1\medskipamount]
\end{tabular}
\end{table}   
\noindent\textbf{Activity 4: Validate EB Safety Requirements} The required input to Activity 4 is the EB Safety Requirements [H].  
These are validated by both review and simulation.
Firstly, the requirements derived for the cloakroom have been reviewed by a safety-critical systems engineering expert to ensure that the specified EB safety requirements for the swarm will deliver its intended safe operation. Secondly, we validated all safety requirements (excepting RQ3.5 from Table~\ref{tab:reqs}) for the cloakroom system using the Gazebo 3D simulator. 
This simulation is an exact replica of the 4m x 4m lab environment used for hardware implementation (see Fig.~\ref{3Dsim}). 
In \textbf{Activity 5}, the artefacts generated in this stage are used to instantiate the EB Safety Requirements Argument Pattern [I].

\subsection{Stage 3: Data Management} \label{framework-stage3}
When designing EB, input data for training an algorithm comes from local sensing of individual agents, both onboard the agent itself and in its local environment. The activities and outputs in this stage take into account the complexities of interactions between multiple agents.

\begin{figure*}[!b]
\centering
\includegraphics[width=0.65\textwidth]{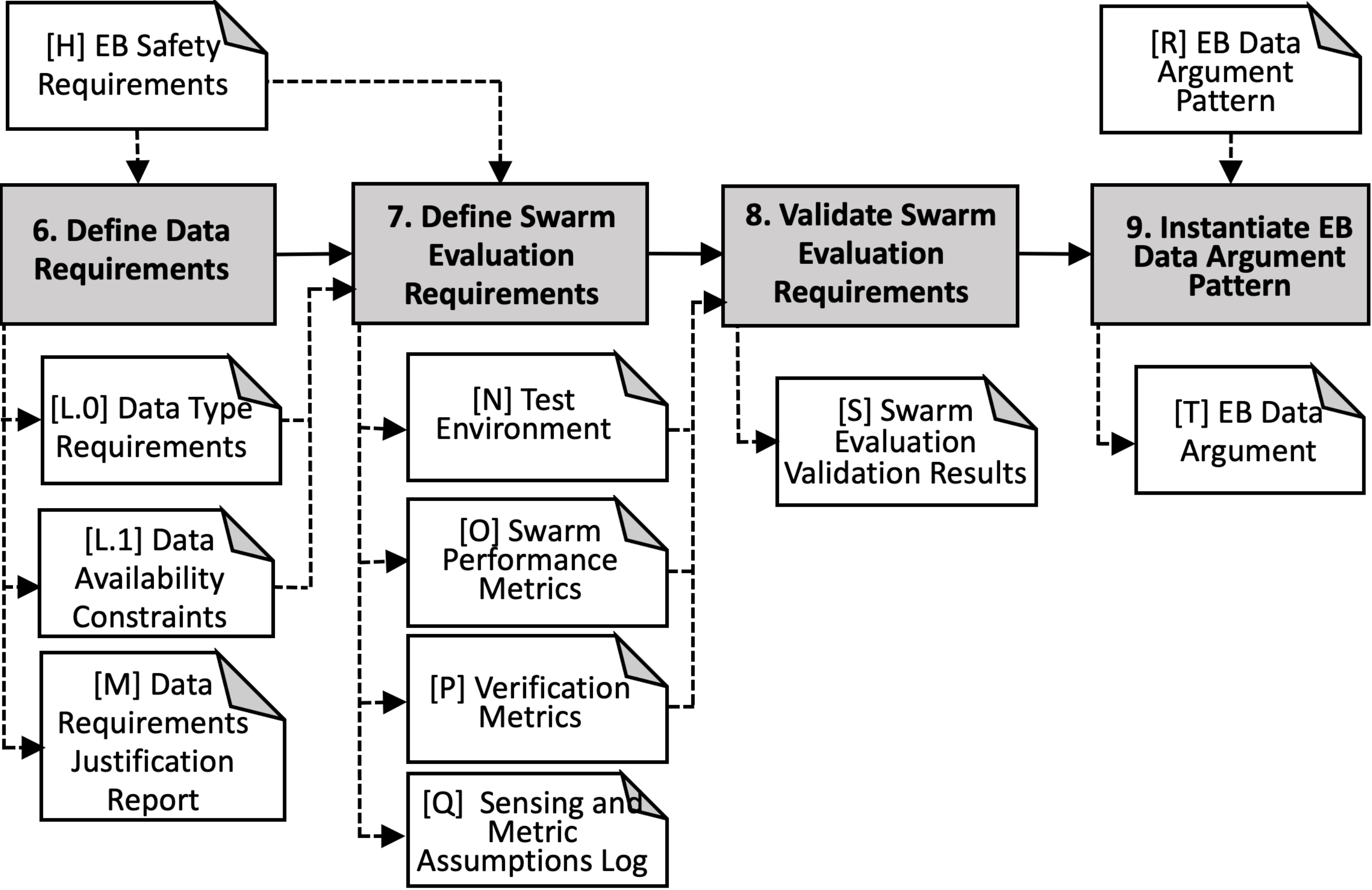}
\caption{Stage 3: The AERoS data management process.}
\label{aeros-stage3}
\end{figure*}

\noindent\textbf{Activity 6. Define Data Requirements} In our adaptation of Activity 6, we take the EB Safety Requirements [H] outlined in Stage 2 as an input (see Fig.~\ref{aeros-stage3}). These safety requirements guide the data requirements in this activity, feeding into the data specification outlined here. We split the data requirement outputs into two multi-agent focused requirements: [L.0] Data Type Requirements and [L.1] Data Availability Constraints.

\emph{[L.0] Data Type Requirements:}
This element focuses on the \emph{relevance}, \emph{completeness}, \emph{accuracy}, and \emph{balance} of the information that will be used to construct the swarm behaviour, and subsequently, to test the EB of the system before deployment. The \emph{relevance} of the data used in the development of the EB specifies the extent to which the test environment must match the intended operating domain of the robot swarm. The \emph{completeness} of the data specifies the conditions under which we test the behaviour, that is, the volume of experiments or tests that will be run, the variety of tests executed, and the diversity of environments expected to be used in the testing process. The aim is to cover a representative sample of conditions for testing. 
\emph{Accuracy} in this context relates to how well the data captures the parameter space defining the performance of the robot swarm. For example, an accurate dataset for what constitutes a delivery in a logistics scenario~\cite{milner2022stochastic} should track the footprint of a deliverable to ensure it is well-positioned in the delivery zone (RQ7.1). 
\emph{Balance} refers to the evenly distributed trials executed in the testing process of the EB algorithm. 
By considering balance, we expect the number of tests conducted for failure modes or environments to be justified, ensuring that there is not an unrealistic bias in testing towards a particular scenario. See Table~\ref{tab:L0_req} for examples of data requirements relating to relevance, completeness, accuracy, and balance.

\begin{table}[!h]
\centering
\caption{Examples of requirements for output [L.0].}
\label{tab:L0_req}
\begin{tabular}{p{0.6cm} p{14cm}}
\textbf{RQ} & \textbf{Relevance Requirements Examples} \\
\hline
5.1 & All simulations \emph{shall} include environments with ranges of incline between 0-20\textdegree.\\
\hline
5.2 & All simulations \emph{shall} be conducted in a dry environment.\\
\hline \\[-1.25\medskipamount]
& \textbf{Completeness Requirements Examples} \\
\hline
6.1 & All simulations \emph{shall} be repeated to include occurrences of faults representative of full communication faults.\\
\hline
6.2 & All simulations \emph{shall} be repeated a sufficient number of times to ensure results are representative of typical use.\\
\hline
6.3 & All simulations \emph{shall} be repeated in multiple environments representative of those expected in real-world use of the system.\\
\hline \\[-1.25\medskipamount]
& \textbf{Accuracy Requirements Examples} \\
\hline
7.1 & All boxes \emph{shall} only be considered `delivered’, if all four of the boxes’ feet are positioned within the delivery zone.\\
\hline
7.2 & All boxes \emph{shall} only be considered `delivered’, once they are no longer in direct contact with a swarm agent.\\
\hline \\[-1.25\medskipamount]
& \textbf{Balance Requirements Examples} \\
\hline
8.1 & All simulations \emph{shall} be repeated so as to obtain representative evaluations for each possible mode of failure (defined under performance, adaptability, and human-safety requirements in Stage 2).\\
\hline
8.2 & All simulations \emph{shall} be repeated equally across all test environments.\\
\hline \\[-1\medskipamount] 
\end{tabular}
\end{table}

\emph{[L.1] Data Availability Constraints:}
With the introduction of multiple agents comes the issue of data availability. Distributed communication is a key feature found in emergent systems. As such, it is crucial to define how much information each agent is expected to hold, how easily data may transfer between agents, and across what range agents should be able to transfer information between one another. Feasible constraints include~\cite{Jones2022}: (i) \emph{storage capacity: }the swarm agents \emph{shall} have a maximum of 2 GB of information stored on board at any time; (ii) \emph{available sensors:} the swarm agents \emph{shall} only have access to environmental data deemed feasibly collectable by radially positioned cameras and laser time-of-flight sensors; (iii) \emph{communication range:} the swarm agents \emph{shall} only have access to other agent data when within communications range of 5 metres; and (iv) \emph{operator feedback:} the swarm agents \emph{shall} only share information with non-agents (e.g.\ operator terminal) when within communications range of 5 metres.

\emph{[M] Data Requirements Justification Report:}
This report is an assessment of the data requirements, providing analysis and explanation for how the requirements and constraints ([L.0] and [L.1]) address the EB Safety Requirements [H].

\noindent\textbf{Activity 7. Define Swarm Evaluation Requirements} Taking the outputs [L.0] and [L.1] from Activity 6, the evaluation requirements consider how the EB of the swarm will be assessed, specifying the testing environment and the metrics to be used to assess the test results. 

\emph{[N] Test Environment:} This takes into consideration the requirements specified in Activity 6, and defines the environment in which the EB will be tested. In most cases this will be multiple simulation environments featuring diverse sets of the terrain, environmental conditions, and obstacle configurations. There may also be instances in which this test environment is specified as a physical environment operating under laboratory conditions, with a hardware system acting as a test bed to observe designed behaviours.

\emph{[O] Swarm Performance Metrics:} This output is used to quantify how well the system is performing. While there may be multiple performance metrics, these metrics should be defined with respect to the primary function of the robot swarm. Metrics that might feature in this output could include: the delivery rate in a logistics scenario, the rate of area coverage in an exploration task, or the response time in disaster scenarios.

\emph{[P] Verification Metrics:} These metrics should be derived from the EB Safety Requirements [H] specified in Stage 2. 
They are intended to be used as the criteria for success within the verification process. 
For example, swarm density, which is used in verifying environmental safety specifications such as RQ3.1, maximum collision force experienced by agents, which could be used to verify that the swarm meets performance requirements such as RQ1.1 and RQ1.2, or the current speed of all agents, a metric relating directly to the human-safety requirements RQ4.1 and RQ4.2. 
Identifying [P] early, ideally during the requirements assurance stage, allows consideration of [P] during the design and development of the swarm to facilitate verification.  

\emph{[Q] Sensing and Metric Assumptions Log:} This log serves as a record of the details and decisions made in Activities 6 and 7. It should contain details of the choices made when producing the Test Environment [N], Swarm Performance Metrics [M], and the Verification Metric [P].

\noindent\textbf{Activity 8. Validate Evaluation Requirements} Taking into account outputs [N], [O], and [P] from Activity 7, this activity aims to validate these components with respect to the requirements specified in Activity 6. Should any discrepancies exist between the data requirements and the evaluation requirements, they should be fully justified and recorded in the output Swarm Evaluation Validation Results [S]. 
The artefacts generated in this stage are used to instantiate the EB Data Argument Pattern [R] in \textbf{Activity 9}.

\subsection{Stage 4: Model Emergent Behaviour} \label{framework-stage4}
In the design of an EB algorithm, the challenge is in selecting behaviours at the individual level of the agents which give rise to the desired EB at the swarm level. 
In our adaptation of AMLAS for the robot swarm,  we step away from the machine learning paradigm to allow consideration for all possible optimisation algorithms which may attain the target EB.

\begin{figure}[!h]
	\centering
	\includegraphics[width=0.5\textwidth]{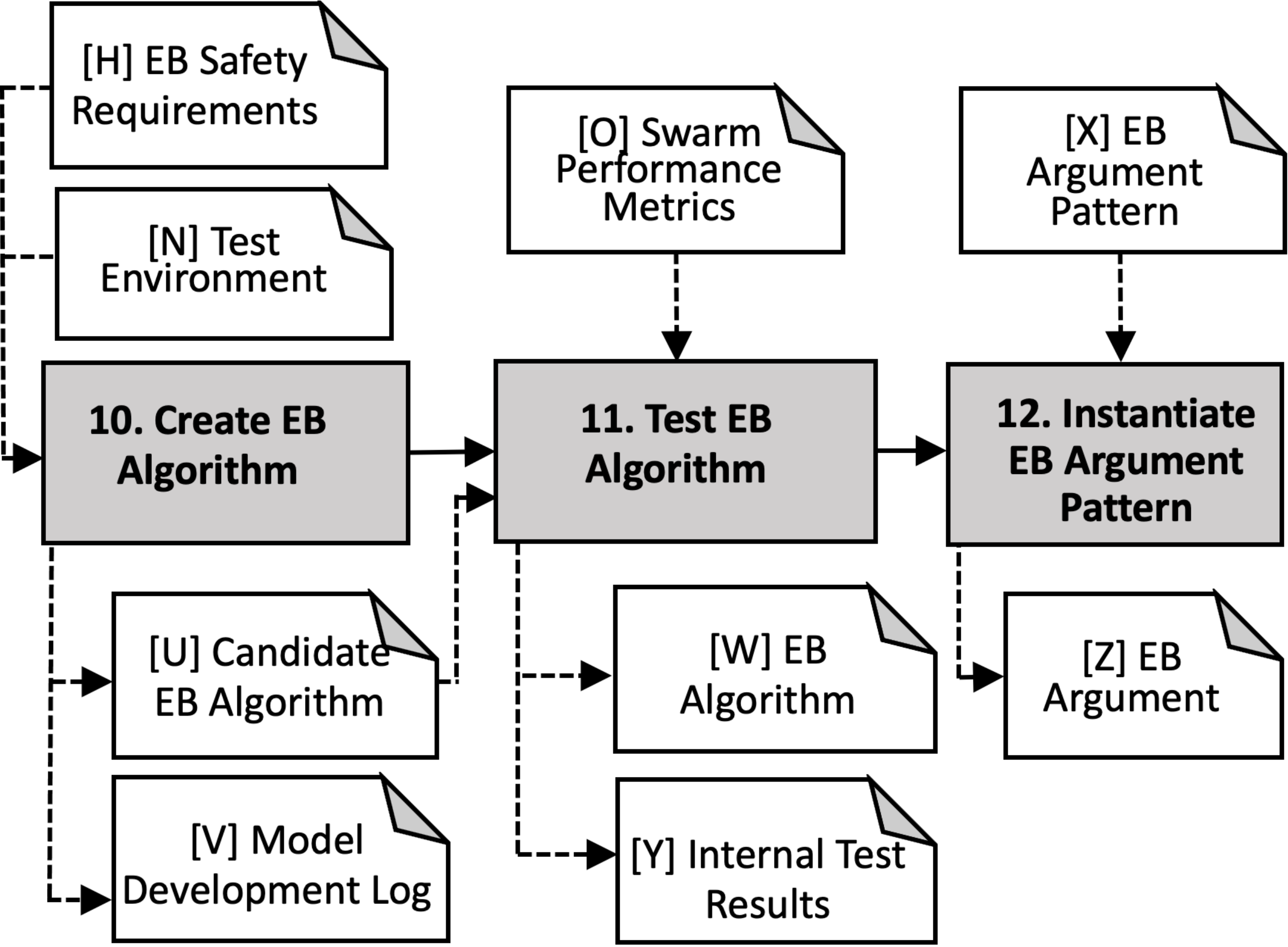}
	\caption{Stage 4: The AERoS model learning process.}
	\label{aeros-stage4}
\end{figure}

\noindent\textbf{Activity 10. Create EB Algorithm} 
This can be nature inspired, hand designed, or evolved from a relatively simple set of instructions for individual behaviour, which takes into account agent-to-agent and environmental interactions~\cite{Jones2018}. These instructions when given to a large number of agents, create a synergistic behaviour for the swarm that is more powerful than the sum of the individual agent's performance. 
The EB algorithm is engineered at the level of the individual agent behaviours for the Test Environment output [N] from Stage 3. The resultant EB must meet the Safety Requirements [H] defined in Stage 2 (see Fig.~\ref{aeros-stage4}). 
In the cloakroom case study, the target EB for the swarm must ensure that items are stored and retrieved by individuals whilst meeting all requirements specified. For example, performance requirements RQ1.1 and RQ1.2 specify an upper bound on the low/high-impact collisions that a swarm shall experience in a given time frame. 
These requirements may be fulfilled by constraining the maximum velocity of individual robots or by ensuring that a robot has one or more sensory devices, such as a camera, enabling it to detect obstacles. 
The key output from this activity is the Candidate EB [U] for testing.

\emph{[V] Model Development Log:} This should log the rationale in the design process of the EB algorithm, in particular how all Safety [H] and Data Type Requirements [L.0] have been met given the Data Availability Constraints [L.1].

\noindent\textbf{Activity 11. Test EB Algorithm} In this activity, the candidate EB will be tested against the Swarm Performance Metrics [O] produced in Stage 3. Testing ensures that the EB performs as desired with respect to the defined metrics and in the case where performance passes accepted thresholds, the EB Algorithm [W] will be produced as the output of the activity. 

\emph{[Y] Internal Test Results:} This output provides a degree of transparency in the testing procedure as the results may be further examined to ensure tests have run correctly. 
In \textbf{Activity 12}, the artefacts generated in this stage are used to instantiate the EB Argument Pattern [X].

\subsection{Stage 5: Model Verification} \label{framework-stage5}
\begin{figure}[!t]
	\centering
	\includegraphics[width=0.4\textwidth]{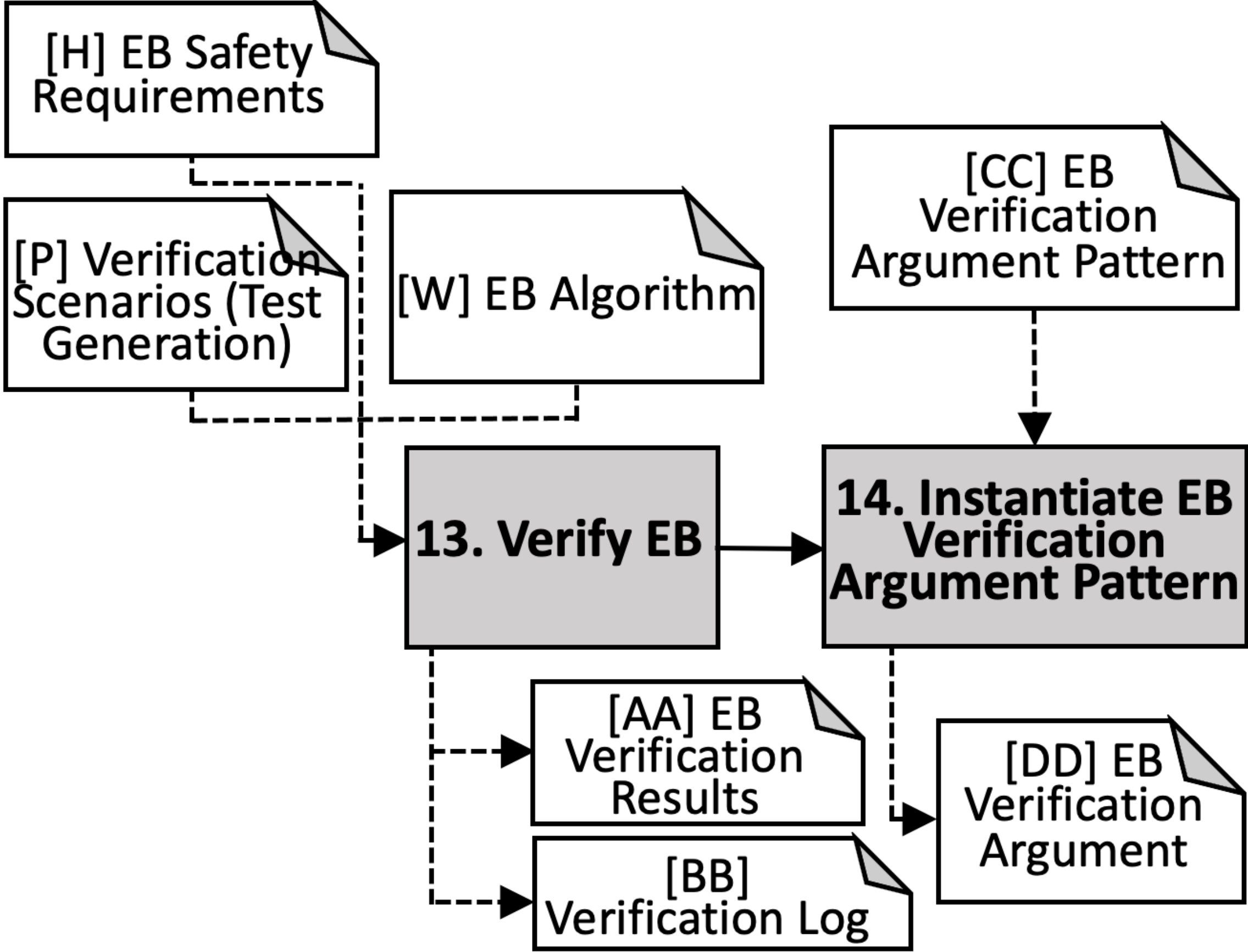}
	\caption{Stage 5: The AERoS verification process.}
	\label{aeros-stage5}
\end{figure}

\noindent\textbf{Activity 13. Verify EB} The inputs to the verification process are the EB Safety Requirements [H], Verification Scenarios (Test Generation) [P], and the EB Algorithm [W] (see Fig.~\ref{aeros-stage5}). 
The verification method and assessment process within that method will be largely determined by the specifics of the safety requirements. Some safety specifications lend themselves towards certain assessment methods due to the scenarios they prescribe.
For example, to assess that the robot swarm meets the requirements for performance given a motor-fault occurrence (see RQ1.3), it may be easier to realise this in physical or simulation-based testing approaches rather than attempting to construct a formal model of robot behaviour given the complex physical dynamics of a faulty wheel.

However, when considering the adaptability requirements, a formal, probabilistic verification technique of the EB Algorithm [W] is more suitable. For example, in RQ2.1, analysis using a probabilistic finite state machine of the swarm behaviour could identify the dwell period within states. Monitors could be used to observe when agents enter a stationary state, for example, \texttt{agent\_velocity=0 $\land $  t\_counter  $\ge$ 100}, and identify if time within that state exceeds some fixed value, and ascertain a probabilistic value to this metric.

\emph{[P] Verification Scenario (Test Generation):} In most cases there will be multiple, valid verification scenarios (test cases) applicable for each of the safety specifications. 
A `good test case' must be \emph{effective} at finding defects, \emph{efficient} in minimising the number of tests required, use resources \emph{economically} and be \emph{robust} to system changes~\cite{Fewster1999}. 
An example of a test case could include a scenario of the swarm in a hazardous environment where too many boxes create an obstacle.

Verification Results [AA] from individual assessments form entries in the Verification Log [BB]. The Verification Log identifies assessments where assurance of the EB Algorithm [W] is acceptable with respect to the Safety Requirements [H] and can be used as a set of evidence for building an assurance case. 
The artefacts generated in this stage are used to instantiate the EB Verification Argument Pattern [CC] in \textbf{Activity 14}.

\subsection{Stage 6: Model Deployment} \label{framework-stage6}
\begin{figure}[!h]
	\centering
	\includegraphics[width=0.55\textwidth]{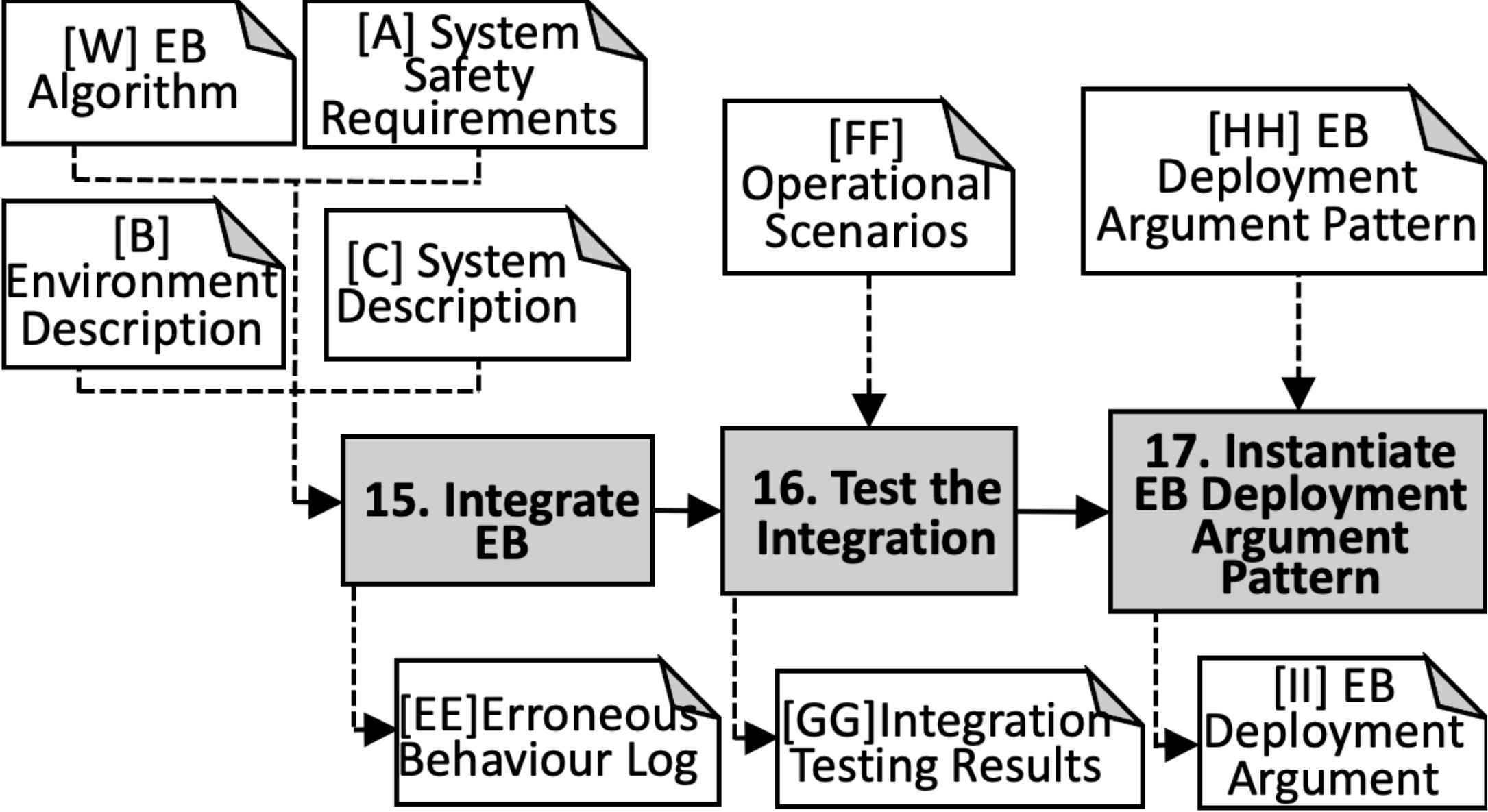}
	\caption{Stage 6: The AERoS model deployment assurance process.}
	\label{aeros-stage6}
\end{figure}

\noindent\textbf{Activity 15. Integrate EB} With the EB verified, the next step is to take the EB Algorithm [W], System Safety Requirements [A], Environment Description [B], and System Description [C] and integrate the EB with the system to be deployed (see Fig.~\ref{aeros-stage6}). 

In this activity, we use the inputs to this stage to inform the implementation of the EB and anticipate errors we might expect in the interactions between agents and the overall EB. Despite the rigorous validation and testing conducted in previous stages, there will still be a gap between the test environment and the intended, everyday-use, deployed scenario. The output, [EE] Erroneous Behaviour Log, captures these anticipated gaps between testing and reality and the differences in behaviour that may surface. 

\noindent\textbf{Activity 16. Test the Integration} Once the initial integration is complete, the physical implementation should undergo additional testing in which the system will be observed in multiple operational scenarios, as specified in [FF].

\emph{[FF] Operational Scenarios:} These scenarios should reflect the environment descriptions specified in [B], offering real-world situations to examine the behaviour of the integrated system. The testing of the integrated system in these true-to-operation environments should be conducted in a safe manner, ensuring that the entire multi-agent system can be shut down in an emergency. 
In the cloakroom, an example of [FF] may take the form of a deployment of agents in a controlled storage area that will not interfere with emergency services.

\emph{[GG] Integration Testing Results:} Results from the integration testing will be reported here, detailing how the system performs against the EB Safety Requirements [H] specified in Stage 2. 
The artefacts generated in this stage are used to instantiate the EB Deployment Argument Pattern [HH] in \textbf{Activity 17}.
\section{Discussion and Future Work} \label{discussion-conclusions}
Using AMLAS \cite{Hawkins2021} as a foundation, we have produced the six-stage development process AERoS. This process acts as guidance for those looking to construct swarm robot systems, particularly those that exhibit emergent behaviour through environmental and agent-to-agent interaction. The stages of AERoS break down the design of these systems to ensure that fundamental safety requirements are adhered to, even in instances of system degradation and compounded failures that should be expected, and managed, in robot swarms. We achieve this with an approach that allows for iteration of and feedback to the previous stages as issues of safety are encountered and investigated. We combine this iteration with repeated specification at each stage, observing the issue of safety through the lens of: data, modelling/behaviour design, verification, and deployment. 

While the iterative nature of AERoS is a key advantage, some limitations have been identified. 
First, the scope of this work has been limited to investigating inherent swarm qualities and the emergent properties that arise from these. 
However, one can expand on this, and consider adaptation of individual robots through techniques such as machine learning (e.g.\ by applying AMLAS).
Second, we can broaden the evaluation by considering additional swarm use cases (e.g.\ monitoring fires in a natural environment, and also a social swarm), and by providing a worked example of the entire AERoS process.

While the focus of the AERoS process is to ensure the safety assurance of EB in swarms, the trustworthiness of an AS can be dependent on many factors other than safety. These include consideration of ethics, and governance and regulation of AS design and operation. 
In future work, we intend to build on Porter et al.’s \cite{Porter2022} Principle-based Ethical Assurance Argument for AI and Autonomous Systems and develop ethics requirements for swarm robots around the ethical principles of beneficence, non-maleficence, respect for autonomy, and justice. 
In addition to ethics requirements, we intend to introduce regulatory requirements into the consideration of AS specification. In particular, we observe the work of Macrae’s~\cite{macrae2021learning} Structural, Organisational, Technological, Epistemic, and Cultural (SOTEC) framework to help us identify sources of socio-technical risk in Autonomous and Intelligent systems. Viewing regulatory requirement analysis from a socio-technical perspective allows us to move away from a purely technical conception of requirements, and helps us design AS that better fit the organisation and operators’ work in which safety considerations are meaningful within the wider system and operational context. The relevance of SOTEC for crafting regulatory requirements for the swarms in the cloakroom as a safety assurance mechanism will be described in a future paper. 

\section*{Acknowledgements} 
The authors would like to thank Alvin Wilby, John Downer, Jonathan Ives, and the AMLAS team for their fruitful comments. The work presented has been supported by the UKRI Trustworthy Autonomous Systems Node in Functionality under Grant EP/V026518/1. I.H.\ is supported by the Assuring Autonomy International Programme at the University of York.

\newpage
\bibliographystyle{unsrt}  
\bibliography{AERoS-Bib.bib}
\end{document}